\documentclass[letterpaper]{article}
\usepackage[letterpaper]{geometry}

\usepackage[utf8]{inputenc}
\usepackage[T1]{fontenc} 
\usepackage[cjk]{kotex}

\usepackage{amsmath}
\usepackage{graphicx}
\usepackage{url}
\usepackage[table]{xcolor}
\usepackage{latexsym}

\usepackage{hyperref}
\definecolor{darkblue}{rgb}{0, 0, 0.5}
\hypersetup{colorlinks=true,citecolor=darkblue, linkcolor=darkblue, urlcolor=darkblue}
\usepackage{tabularx}
\usepackage{datetime}
\usepackage{arydshln}

\usepackage{setspace}
\usepackage{lineno}

\usepackage{rotating}

\usepackage{booktabs}

\usepackage[authoryear,round]{natbib}
\usepackage{algorithmic,algorithm}
\renewcommand{\algorithmiccomment}[1]{\bgroup\hfill$\triangleright$~#1\egroup}

\usepackage[linguistics]{forest} 
\usepackage{synttree}
\usepackage{subcaption}
\usepackage{tikz-dependency}

\usepackage{langsci-gb4e}

\title{\textbf{
Lexically conditioned realization ambiguity in Korean predicate morphology\thanks{Corresponding author: \url{jungyeul@kaist.ac.kr}}
}}

\author{ 
Wonjun Oh, KyungTae Lim, and Jungyeul Park\\
Korea Advanced Institute of Science \& Technology, South Korea
}
\date{ 
}

\begin{document}

\maketitle

\begin{abstract}
This paper examines Korean surface realization as distinct from morphological analysis. It asks whether a sequence of canonical morphemes and grammatical category labels uniquely determines the corresponding surface form. The answer is negative for a restricted but theoretically revealing class of Korean predicates. In these cases, formally identical or near-identical stem-ending configurations yield different outputs depending on lexical identity and realization class membership. We analyze this phenomenon as \textit{homonymy with inflectional divergence}, focusing on regular versus ㄷ \textit{digeut} irregular pairs, regular versus ㅂ \textit{bieup} irregular pairs, and 르 \textit{reu} irregular versus 러 \textit{reo} irregular pairs. These cases show that stem shape and ending alone do not always determine surface realization. Instead, lexical meaning, subcategorization, and semantic role structure help identify the intended predicate; the predicate determines the realization class; and the realization class determines the surface form. Korean realization thus reveals a limit of bare morphological representation.
\end{abstract}

\tableofcontents

\doublespacing

\section{Introduction}

In Korean, eojeol is not a primitive grammatical unit in the sense of a minimal morphological word \citep{lim-2024-terms}.\footnote{Eojeol is the orthographic spacing unit in Korean, typically consisting of a lexical stem combined with one or more functional morphemes, such as case markers or verbal endings; it serves as the basic unit of written word segmentation.} Rather, it is an orthographic unit \citep{kim-2015-establishment} within which lexical and functional material may be combined. A single eojeol may contain a nominal or predicative base together with case markers, derivational elements, honorific markers, tense and aspect morphology, clause typing morphology, or sentence final material. Korean morphological analysis therefore decomposes the eojeol into a sequence of morphemes and assigns each segment a grammatical category. This analytic move shifts attention from the surface cohesion of the written unit to the internal configuration of stems and formatives.

Once the eojeol is represented in this segmented way, however, an inverse question arises. Given a sequence of canonical morphemes together with part-of-speech labels, under what conditions does this representation determine a unique surface form? Put differently, what is the relation between morphological analysis and morphological realization in Korean? The present paper takes this question as its starting point. Its concern is not segmentation itself, but the reverse mapping from an analyzed representation to the overt form that appears in actual usage \citep{anderson-1992-a-morphous,stump-2001-inflectional,stump-2016-inflectional}.

This reverse mapping cannot be reduced to simple concatenation. Such a view would treat the output of analysis as if it were merely the surface string broken into parts, so that realization could be obtained by restoring adjacency. Korean morphology does not permit so simple a conception \citep{song-1995-irregularities}. In many environments, especially in predicate formation, the relation between a stem and a following ending is mediated by alternation, contraction, deletion, vowel adjustment, consonantal modification, or other morphophonological effects that are not recoverable from linear juxtaposition alone \citep{kim-2015-conjugated}. The surface eojeol is therefore not a mechanical sum of its segmented components, but the output of a realization process governed by language-specific conditions on how morphemes are overtly expressed.

The limits of this process become especially clear in so-called irregular conjugation. Traditional descriptions distinguish regular and irregular predicate classes, but this opposition conceals a more complex landscape. Korean irregularity is not a single homogeneous phenomenon, but a graded domain whose instances differ in distribution, predictability, and relation to corresponding regular forms \citep{park-2004-irregularity}. More importantly for the present paper, some formally identical or near-identical predicate stems diverge in realization because they belong to different lexical items and inflectional classes.\footnote{By \textit{near-identical predicates}, we mean predicates that contain the same lexical base but occur in morphologically expanded forms, often through prefixation or compounding. For example, 누르다 \textit{nureuda} (`press'), 억누르다 \textit{eongnureuda} (`suppress'), and 짓누르다 \textit{jinnureuda} (`crush down, weigh down') share the base 누르다 \textit{nureuda} and preserve the relevant realization behavior.} The problem is therefore not simply that Korean has irregular forms, {nor simply that Korean has homonymous predicates. Rather, it is that homonymous or near-homonymous predicates may belong to different realization classes, so that} the same analyzed stem-ending configuration may fail to determine a unique surface output unless the intended lexeme and its realization class are identified.

We call this phenomenon \textit{homonymy with inflectional divergence}. In such cases, homonymous or near-homonymous predicates share a stem shape but differ in realization behavior. The contrast is not determined by stem shape and ending alone, nor by semantic roles directly. Rather, argument structure and lexical meaning help identify the intended lexeme; the lexeme determines the relevant realization class; and that realization class determines the surface form. The relevant direction of explanation is therefore:
\[
\text{argument frame / semantic role pattern}
\rightarrow
\text{lexical identity}
\rightarrow
\text{realization class}
\rightarrow
\text{surface form}.
\]
This architecture preserves a morphology-first account of realization while showing why lexical semantic information is necessary for disambiguating the intended predicate.

This problem has rarely been treated as a theoretical issue in its own right. Korean linguistics has long examined conjugational classes, stem alternations, morphophonological adjustment, and the status of irregular forms. Less explicitly addressed is the question of realization when posed from the perspective of an already segmented analysis: given a morphologically decomposed representation, what principles allow the grammar to recover the correct eojeol, and where does recovery fail unless further lexical information is supplied? {From this perspective, homonymy with inflectional divergence is not merely a lexicographic observation about pairs of homonymous predicates. It is a test case for the sufficiency of morphological representation in surface realization.} This question lies at the intersection of several central issues in morphology, including the relation between analysis and surface realization, the division of labor between lexicon and rule, the status of gradient irregularity, and the extent to which surface form is predictable from morphosyntactic representation.

The present paper addresses this gap by examining Korean surface form generation as a problem of morphological realization. We begin with the standard output of morphological analysis, namely a sequence of morphemes annotated for grammatical category, and ask whether such a representation is sufficient to determine the overt eojeol. Our central claim is that it is not sufficient in any general sense. While many nominal forms approach transparent compositional realization, predicate forms reveal a more intricate interaction among stem identity, realization class membership, and morphophonological conditioning. By bringing together the relevant ambiguity classes, including regular versus ㄷ \textit{digeut} irregular, regular versus ㅂ \textit{bieup} irregular, and 르 \textit{reu} irregular versus 러 \textit{reo} irregular pairs, the paper argues that Korean predicate realization provides a revealing domain for studying how lexical identity and inflectional behavior are connected. {The contribution is therefore not the observation that individual irregular predicates exist, but the formulation of a restricted class of Korean predicates as a realization ambiguity problem: a morphologically analyzed input may remain compatible with more than one surface form until lexical identity and realization class membership are fixed.} {An appendix provides corpus support from the KLUE dependency treebank and reports a surface reconstruction experiment that operationalizes the realization problem discussed here.}

\section{Morphological analysis and surface realization in Korean}

Korean morphological analysis represents an eojeol as an ordered sequence of morphemes, each associated with a grammatical category. A nominal eojeol may consist of a lexical base followed by a case marker or auxiliary particle, while a predicate eojeol may contain a lexical stem followed by endings expressing tense, honorification, clause type, speech level, or nominal modification. Such an analysis makes the internal composition of the surface form explicit, but it does not by itself specify all conditions required for overt realization.

The relation between analyzed structure and surface form varies across morphological domains. In many nominal combinations, the mapping is relatively transparent. When a noun combines with a postposition, the segmental material of both elements is commonly preserved with only limited boundary adjustment. A representation such as \{학교+에\} may therefore correspond straightforwardly to 학교에 \textit{hakgyoe} (`at school'). In such cases, the analyzed representation is not identical to the surface string, but the distance between the two remains small.

Predicate morphology presents a different situation. When a verbal or adjectival stem combines with an ending, the resulting form may depart from linear concatenation. A representation such as \{가+아\} is realized not as *가아 \textit{ga-a}, but as 가 \textit{ga} (`go'), reflecting contraction. Likewise, \{오+아\} is realized as 와 \textit{wa} (`come'), not as *오아 \textit{o-a}. These examples are regular in the sense that their outcomes are predictable once the relevant morphophonological conditions are known, but they already show that realization is not mere adjacency restoration.

The distance becomes greater when consonantal alternation and stem allomorphy are involved. Predicate stems that are formally similar in canonical representation may pattern differently before the same ending. A stem ending in ㄷ \textit{digeut} may preserve that consonant in one lexical item but alternate with ㄹ \textit{rieul} in another. A stem ending in ㅂ \textit{bieup} may retain ㅂ \textit{bieup} in one item but alternate with 우 \textit{u} or 오 \textit{o} in another. A stem in 르 \textit{reu} may participate in one realization class for one lexeme and in another realization class for a distinct lexeme. {These are not merely cases in which a phonological environment licenses different outputs. They are cases in which homonymous or near-homonymous predicates may differ in inflectional behavior.} In such cases, the analyzed sequence identifies the morphemes involved, but not necessarily the realization class to which the predicate belongs.

For present purposes, three relations between analysis and realization should be distinguished. First, realization may be nearly transparent, with the surface form largely preserving the segmental content and order of the analyzed morphemes, as in \{먹+었+다\} $\rightarrow$ 먹었다 \textit{meogeotda} (`ate'). Second, realization may be non-transparent but locally predictable, once the lexical item and the relevant morphophonological conditions are known, as in \{나서+었+다\} $\rightarrow$ 나섰다 \textit{naseotda} (`stepped forward' or `came forward'). Third, realization may be underdetermined by the analyzed sequence because formally identical or near-identical predicate representations remain compatible with distinct surface outputs, as in \{걷+었+다\}, which may yield 걸었다 \textit{georeotda} (`walked') or 걷었다 \textit{geodeotda} (`cleared away' or `collected'), depending on the intended lexeme. {This third relation is the domain of homonymy with inflectional divergence: the same analyzed stem-ending configuration is compatible with more than one realization because distinct lexical items with the same or similar stem shape belong to different realization classes.}

It is this third relation that motivates the analysis developed below. In such cases, the analyzed morpheme sequence and general grammatical rules do not suffice to restore the surface form. The missing step is lexical identification: the grammar must determine which lexeme is intended, since the realization class is associated with the lexeme rather than with the stem shape alone. {Argument structure and lexical meaning can help identify the intended lexeme, but they do not directly determine the surface form; the lexeme determines the realization class, and the realization class determines the output.} The following section develops this point through Korean irregular conjugation, where lexically conditioned realization classes make the limits of formal predictability especially visible.

\section{Irregular conjugation and the limits of formal predictability}

Korean predicate morphology includes a range of patterns traditionally described as irregular conjugation. These include stem-changing types such as ㄷ \textit{digeut}, ㅂ \textit{bieup}, ㅅ \textit{siot}, 르 \textit{reu}, and 우 \textit{u}, ending-changing types such as 러 \textit{reo}, 여 \textit{yeo}, and 오 \textit{o}, and mixed patterns such as ㅎ \textit{hieut}. Some alternations that are superficially similar, notably 으 \textit{eu} and ㄹ \textit{rieul} related patterns, are often treated in school grammar as regular rather than genuinely irregular. The descriptive landscape is therefore broader than a simple opposition between regular and irregular forms.

For the present paper, the relevant point is not the full taxonomy of Korean conjugation, but what irregular conjugation shows about realization classes. Many alternations apply only to predicates that belong to a particular inflectional class. The morphophonological environment may license an alternation, but it does not by itself determine whether the alternation applies. That determination depends on the lexical item. {In this sense, Korean irregular conjugation is not only a set of local stem-ending adjustments, but also a system of lexically indexed realization classes.}

This can be seen clearly in ㄷ \textit{digeut} irregular predicates. In this class, stem-final ㄷ \textit{digeut} changes to ㄹ \textit{rieul} before vowel-initial endings. Thus \{듣+어\} yields 들어 \textit{deureo} (`hear'), and \{묻+어\} yields 물어 \textit{mureo} (`ask'). Yet not every stem ending in ㄷ \textit{digeut} behaves in this way. A regular predicate such as \{닫+아\} surfaces as 닫아 \textit{dada} (`close'), not *달아 \textit{dal-a}, unless a different lexical item is involved. The final consonant and following ending therefore define the possible site of alternation, but class membership determines whether the alternation is realized. {The same surface stem shape may thus be compatible with more than one realization pattern only because it corresponds to more than one lexical item.}

A similar contrast appears with ㅂ \textit{bieup} irregular predicates. In many items, stem-final ㅂ \textit{bieup} alternates with 우 \textit{u} or 오 \textit{o} before certain endings, as in \{굽+어\} $\rightarrow$ 구워 \textit{guwo} (`roast'), \{줍+어\} $\rightarrow$ 주워 \textit{juwo} (`pick up'), and \{곱+아\} $\rightarrow$ 고와 \textit{gowa} (`be lovely'). Other predicates with stem-final ㅂ \textit{bieup}, however, preserve the consonant. Here again, phonological shape identifies a possible environment, but the realization pattern is lexically specified. {The alternation is therefore predictable only after the predicate has been assigned to the appropriate lexical realization class.}

The same point extends to 르 \textit{reu} and 러 \textit{reo} related patterns. In 르 \textit{reu} irregular predicates, the stem-final syllable 르 \textit{reu} is reduced, and the following ending is realized as 라 \textit{ra} or 러 \textit{reo}, as in \{누르+어\} $\rightarrow$ 눌러 \textit{nulleo} (`press') and \{모르+아\} $\rightarrow$ 몰라 \textit{molla} (`not know'). By contrast, 러 \textit{reo} irregular predicates involve a different relation between stem and ending, as in \{이르+어\} $\rightarrow$ 이르러 \textit{ireureo} (`reach') and \{푸르+어\} $\rightarrow$ 푸르러 \textit{pureureo} (`be blue'). These patterns are important because formally similar stems may belong to distinct realization classes, and those classes may correlate with lexical meaning. {The correlation is not a direct mapping from meaning to form; rather, lexical meaning helps distinguish the relevant lexeme, and the lexeme carries the realization class.}

Other irregular patterns confirm that Korean predicate realization is organized by multiple lexically conditioned classes rather than by a single exceptional mechanism.
In ㅅ \textit{siot} irregular predicates, stem-final ㅅ \textit{siot} disappears before vowel-initial endings, as in \{짓+어\} $\rightarrow$ 지어 \textit{jieo} (`build'). In the unique 우 \textit{u} irregular item 푸다 \textit{puda} (`scoop'), stem-final 우 \textit{u} disappears before \{-어\}, yielding 퍼 \textit{peo}. In 여 \textit{yeo} irregular forms, represented most prominently by 하다 \textit{hada} (`do') and 하다 compounds, \{-아\} appears as \{-여\}, yielding 하여 \textit{hayeo} and the contracted form 해 \textit{hae}. In 오 \textit{o} irregular forms, the imperative ending \{-아라/-어라\} is replaced by \{-오\} in the unique item 달다 \textit{dalda} (`ask for, request'), yielding 다오 \textit{dao}. In ㅎ \textit{hieut} irregular adjectives, stem-final ㅎ \textit{hieut} disappears before certain endings and combines with \{-아/-어\} to produce forms such as 까매 \textit{kkamae} (`be black'), 노래 \textit{norae} (`be yellow'), and 하얘 \textit{hayyae} (`be white').

Table~\ref{tab:korean_irregular_conjugation_overview} summarizes the major Korean conjugation types relevant to the present discussion, including stem-changing, ending-changing, mixed, and regularly classified alternation patterns.

\begin{table}[!ht]
\centering
\footnotesize
\begin{tabular}{p{.2\textwidth} p{.15\textwidth} p{.55\textwidth}}
\toprule
\textbf{Type} & \textbf{Classification} & \textbf{Description and examples} \\
\midrule

ㄷ \textit{digeut} irregular & stem alternation &
Stem-final ㄷ \textit{digeut} changes to ㄹ \textit{rieul} before a vowel-initial ending:
듣다 \textit{deutda} (`hear') $\rightarrow$ 들어 \textit{deureo}, 묻다 \textit{mutda} (`ask') $\rightarrow$ 물어 \textit{mureo} \\
\midrule

ㅂ \textit{bieup} irregular & stem alternation &
Stem-final ㅂ \textit{bieup} is realized as 우 \textit{u} or 오 \textit{o}:
굽다 \textit{gupda} (`roast') $\rightarrow$ 구워 \textit{guwo}, 줍다 \textit{jupda} (`pick up') $\rightarrow$ 주워 \textit{juwo}, 곱다 \textit{gopda} (`be lovely') $\rightarrow$ 고와 \textit{gowa} \\
\midrule

ㅅ \textit{siot} irregular & stem alternation &
Stem-final ㅅ \textit{siot} is deleted before a vowel-initial ending:
짓다 \textit{jitda} (`build') $\rightarrow$ 지어 \textit{jieo}, 낫다 \textit{natda} (`heal') $\rightarrow$ 나아 \textit{naa} \\
\midrule

르 \textit{reu} irregular & stem alternation &
Stem-final 르 \textit{reu} is reduced, and the ending is realized as 라 \textit{ra} or 러 \textit{reo}:
누르다 \textit{nureuda} (`press') $\rightarrow$ 눌러 \textit{nulleo}, 모르다 \textit{moreuda} (`not know') $\rightarrow$ 몰라 \textit{molla} \\
\midrule

우 \textit{u} irregular & stem alternation &
Stem-final 우 \textit{u} is deleted before \{-어\}:
푸다 \textit{puda} (`scoop') $\rightarrow$ 퍼 \textit{peo} \\
\midrule

러 \textit{reo} irregular & ending alternation &
The \{-어\} portion of the endings \{-어/-어서\} changes to \{-러\}:
이르다 \textit{ireuda} (`reach') $\rightarrow$ 이르러 \textit{ireureo}, 푸르다 \textit{pureuda} (`be blue') $\rightarrow$ 푸르러 \textit{pureureo} \\
\midrule

여 \textit{yeo} irregular & ending alternation &
The ending \{-아\} changes to \{-여\}:
하다 \textit{hada} (`do') $\rightarrow$ 하여 \textit{hayeo} $\rightarrow$ 해 \textit{hae}, 공부하다 \textit{gongbuhada} (`study') $\rightarrow$ 공부하여 \textit{gongbuhayeo} $\rightarrow$ 공부해 \textit{gongbuhae} \\
\midrule

오 \textit{o} irregular & ending alternation &
The imperative ending \{-아라/-어라\} changes to \{-오\}:
달다 \textit{dalda} (`ask for, request') $\rightarrow$ 다오 \textit{dao} \\
\midrule

ㅎ \textit{hieut} irregular & stem plus ending alternation &
In some adjectives, stem-final ㅎ \textit{hieut} is deleted or combines differently with the following ending:
까맣다 \textit{kkamata} (`be black') $\rightarrow$ 까매 \textit{kkamae}, 노랗다 \textit{norata} (`be yellow') $\rightarrow$ 노래 \textit{norae}, 하얗다 \textit{hayata} (`be white') $\rightarrow$ 하얘 \textit{hayyae} \\
\midrule

으 \textit{eu} regular & classified as regular conjugation &
Stem-final 으 \textit{eu} is deleted before \{-아/-어\}, but in school grammar this pattern is usually treated as regular rather than irregular:
크다 \textit{keuda} (`be big') $\rightarrow$ 커 \textit{keo}, 쓰다 \textit{sseuda} (`write') $\rightarrow$ 써 \textit{sseo} \\
\midrule

ㄹ \textit{rieul} regular & classified as regular conjugation &
Stem-final ㄹ \textit{rieul} is deleted before ㄴ \textit{nieun}, ㄹ \textit{rieul}, ㅂ \textit{bieup}, 오 \textit{o}, and 시 \textit{si} (`honorific'), but in school grammar this pattern is usually treated as regular rather than irregular:
갈다 \textit{galda} (`sharpen') $\rightarrow$ 가니 \textit{gani}, 가오 \textit{gao}, 가는 \textit{ganeun} \\

\bottomrule
\end{tabular}
\caption{Overview of major Korean irregular conjugation types}
\label{tab:korean_irregular_conjugation_overview}
\end{table}

Irregular conjugation therefore provides the immediate background for the stronger ambiguity cases examined in the next section. Once a predicate is identified as a member of a particular realization class, many alternations are systematic. The harder cases arise when the relevant class membership is not recoverable from stem shape and ending alone because homonymous or near-homonymous predicates belong to different realization classes. {These cases go beyond the ordinary statement that some predicates are irregular: they show that a morphologically analyzed stem-ending configuration may remain compatible with more than one surface output until lexical identity is fixed.}

{This perspective differs from a traditional description of irregular conjugation in one crucial respect. Traditional descriptions typically state the form of an alternation for a given class: ㄷ \textit{digeut} irregular predicates change ㄷ \textit{digeut} to ㄹ \textit{rieul} before vowel-initial endings, ㅂ \textit{bieup} irregular predicates realize final ㅂ \textit{bieup} as 우 \textit{u} or 오 \textit{o}, and so on. Such descriptions are necessary, but they presuppose that the predicate has already been assigned to the relevant class. The present paper instead asks what happens when the analyzed input itself does not determine that assignment. In cases of homonymy with inflectional divergence, the problem is not how an irregular class is realized once selected, but how the grammar identifies which lexeme, and therefore which realization class, is involved.}

These are the cases of homonymy with inflectional divergence to which we now turn.

\section{Ambiguity in surface realization and lexical disambiguation}

The strongest limit on formal predictability arises when predicates with the same or nearly the same canonical stem shape belong to different realization classes. In such cases, the issue is not simply whether a predicate is irregular in the abstract. Rather, the same stem-ending configuration may yield different surface forms because it instantiates different lexemes. Korean predicate realization is therefore shaped not only by morphophonological alternation, but also by lexical differentiation within the verbal and adjectival lexicon.

This is the phenomenon introduced above as \textit{homonymy with inflectional divergence}. It is clearest in pairs such as \{걷+어\} \textit{geot+eo}, which yields 걷어 \textit{geodeo} when the predicate means `clear away' or `lift and gather', but 걸어 \textit{georeo} when it means `walk'. Similar contrasts recur across the major classes examined in this paper. In 묻다 \textit{mutda} (`bury' or `ask'), 닫다 \textit{datda} (`close' or `run'), and related compounds such as 되묻다 \textit{doemutda} (`ask back'), 치걷다 \textit{chigeotda} (`roll up'), and 파묻다 \textit{pamutda} (`bury'), one lexical item remains regular while another shows ㄷ \textit{digeut} irregular realization. In 곱다 \textit{gopda} (`be lovely' or `turn out unfavorably') and 굽다 \textit{gupda} (`roast' or `bend'), the contrast involves regular versus ㅂ \textit{bieup} irregular realization. In 누르다 \textit{nureuda} (`press' or `be yellowish') and 이르다 \textit{ireuda} (`say' or `reach'), {distinct lexemes with related or identical stem shapes are assigned to} 르 \textit{reu} irregular and 러 \textit{reo} irregular realization {classes}.

These cases are more revealing than irregularity considered in general. Ordinary descriptions of irregular conjugation state which alternation applies once the relevant predicate class is already known. The present cases show that this prior classification may itself remain unresolved. The ambiguity arises because homonymous or near-homonymous predicates are distributed across different inflectional classes. {They therefore constitute a realization ambiguity problem, not merely a list of homonymous dictionary entries.}

The ambiguity classes examined here fall into three recurrent types: regular versus ㄷ \textit{digeut} irregular, regular versus ㅂ \textit{bieup} irregular, and 르 \textit{reu} irregular versus 러 \textit{reo} irregular. Across these types, formal similarity defines only the site at which an alternation could occur. It does not determine whether that alternation is realized. What must be identified is the intended lexeme, since realization class membership is associated with predicates as lexemes rather than derived automatically from their segmental shape. {The direction of explanation is therefore from lexical identification to realization class, not from semantic role structure directly to surface form.}

This point bears directly on {lexical} disambiguation. The missing information is not random, nor is it supplied by semantic roles in a direct way. Rather, lexical meaning and argument structure help identify the intended predicate, and the predicate in turn determines the relevant realization class. The following subsections make this relation explicit by drawing on the subcategorization-frame information provided in the Sejong dictionary. These frames are used not as independent determinants of irregularity, but as evidence for the lexical distinctions that underlie different realization classes. 
Table~\ref{tab:homonym_conjugation} presents the core ambiguity classes.\footnote{{The table provides an exhaustive list of the relevant ambiguity classes identified in the \textit{Standard Korean Language Dictionary} (\textit{Pyojun Gugeo Daesajeon}).}}
The subsections that follow show how these lexical contrasts are reflected in subcategorization frames and semantic role structure.

\begin{table}[!ht]
\centering
\footnotesize

\begin{tabular}{p{.2\textwidth} p{.15\textwidth} p{.2\textwidth} p{.3\textwidth}}
\toprule
\textbf{Predicate} & \textbf{Meaning} & \textbf{Class} & \textbf{Realization examples} \\
\midrule

\multicolumn{4}{l}{\cellcolor{gray!25} Regular vs. ㄷ \textit{digeut} irregular} \\

걷다 \textit{geotda} (03) &
`clear away' &
regular &
걷어 \textit{geodeo}, 걷으니 \textit{geodeuni} \\

걷다 \textit{geotda} (02) &
`walk' &
ㄷ \textit{digeut} irregular &
걸어 \textit{georeo}, 걸으니 \textit{georeuni} \\
\hline

닫다 \textit{datda} (02) &
`close' &
regular &
닫아 \textit{dada}, 닫으니 \textit{dadeuni} \\

닫다 \textit{datda} (01) &
`run' &
ㄷ \textit{digeut} irregular &
달아 \textit{dara}, 달으니 \textit{dareuni} \\
\hline

묻다 \textit{mutda} (02) &
`bury' &
regular &
묻어 \textit{mudeo}, 묻으니 \textit{mudeuni} \\

묻다 \textit{mutda} (03) &
`ask' &
ㄷ \textit{digeut} irregular &
물어 \textit{mureo}, 물으니 \textit{mureuni} \\

\midrule

\multicolumn{4}{l}{\cellcolor{gray!25} Regular vs. ㅂ \textit{bieup} irregular} \\

곱다 \textit{gopda} (01) &
`suffer a loss' &
regular &
곱아 \textit{goba}, 곱으니 \textit{gobeuni} \\

곱다 \textit{gopda} (02) &
`be beautiful' &
ㅂ \textit{bieup} irregular &
고와 \textit{gowa}, 고우니 \textit{gouni} \\
\hline

굽다 \textit{gupda} (02) &
`bend' &
regular &
굽어 \textit{gubeo}, 굽으니 \textit{gubeuni} \\

굽다 \textit{gupda} (01) &
`roast' &
ㅂ \textit{bieup} irregular &
구워 \textit{guwo}, 구우니 \textit{guuni} \\

\midrule

\multicolumn{4}{l}{\cellcolor{gray!25} 르 \textit{reu} irregular vs. 러 \textit{reo} irregular} \\

누르다 \textit{nureuda} (01) &
`press' &
르 \textit{reu} irregular &
눌러 \textit{nulleo}, 누르니 \textit{nureuni} \\

누르다 \textit{nureuda} (02) &
`be yellowish' &
러 \textit{reo} irregular &
누르러 \textit{nureureo}, 누르니 \textit{nureuni} \\
\hline

이르다 \textit{ireuda} (02) &
`say' &
르 \textit{reu} irregular &
일러 \textit{illeo}, 이르니 \textit{ireuni} \\

이르다 \textit{ireuda} (01) &
`reach' &
러 \textit{reo} irregular &
이르러 \textit{ireureo}, 이르니 \textit{ireuni} \\

\bottomrule
\end{tabular}

\vspace{0.2cm}
\raggedright
\scriptsize
\textit{Note.} For 르 \textit{reu} irregular and 러 \textit{reo} irregular predicates, \citet{ha-2012-verbal} argues that different allomorphic patterns arose under a homonymy avoidance constraint, preventing forms with distinct lexical meanings from becoming identical during conjugation. For ㅂ \textit{bieup} irregular predicates, \citet{cho-2009-study} relates the modern ㅂ \textit{bieup}/우 \textit{u} or 오 \textit{o} alternation to a historical weakening process in which the Middle Korean labial fricative ㅸ changed into the glide \textit{w}.

\caption{Comparison of realization patterns across lexically distinct homonymous predicates}
\label{tab:homonym_conjugation}
\end{table}

\subsection{Regular versus ㄷ \textit{digeut} irregular pairs}

Regular versus ㄷ \textit{digeut} irregular pairs show that the mapping from a canonical stem-ending sequence to a surface form may be lexically underdetermined. The relevant predicates have identical citation forms, but their regular and irregular realizations are associated with different lexical meanings and different argument frames. The contrast between 걷어 \textit{geodeo} (`clear away' or `lift and gather') and 걸어 \textit{georeo} (`walk'), 닫아 \textit{dada} (`close') and 달아 \textit{dara} (`run'), and 묻어 \textit{mudeo} (`bury') and 물어 \textit{mureo} (`ask') therefore illustrates homonymy with inflectional divergence: what appears as a single stem shape corresponds to distinct predicates whose realization classes must be determined lexically. {The argument-structural contrasts discussed below are important because they help identify the intended lexical entry; they are not themselves rules that derive the irregular surface form.}

\paragraph{걷다 \textit{geotda}}
The verb 걷다 \textit{geotda} shows a split between two lexical entries sharing the same headword but belonging to different realization classes. The ㄷ \textit{digeut} irregular entry is realized as 걸어 \textit{georeo} before vowel-initial endings and denotes self motion, as in `walk'. Its basic frame is intransitive, as in (\ref{geotda-1}), with $X$ realized as a nominative \textsc{agent}.

\begin{exe}
\ex \label{geotda-1}
    \glll $\{X=N_0\}$-이 $V$\\
          ${\color{white}\{X=N_0\}}$-\textit{i} \\
          ${\color{white}\{X=N_0\}}$-\textsc{nom} \\
\end{exe}

\noindent The same irregular entry also licenses an accusative-marked argument, as in (\ref{geotda-2}). This argument is not an affected object. In the literal motion reading, $Y$ denotes a traversed \textsc{location} or \textsc{path}, as in walking along a street or trail. In extended readings such as `live' or `pass through', $Y$ denotes an abstract \textsc{theme}, such as a path of hardship, a field of activity, or a professional course.

\begin{exe}
\ex \label{geotda-2}
    \glll $\{X=N_0\}$-이 $\{Y=N_1\}$-을 $V$\\
          ${\color{white}\{X=N_0\}}$-\textit{i} ${\color{white}\{Y=N_1\}}$-\textit{eul} \\
          ${\color{white}\{X=N_0\}}$-\textsc{nom} ${\color{white}\{Y=N_1\}}$-\textsc{acc} \\
\end{exe}

\noindent The regular entry, by contrast, is realized as 걷어 \textit{geodeo}. It covers several related transitive meanings, including `roll up', `fold up', `remove', and `finish'. These senses share the frame in (\ref{geotda-3}), where $X$ is an \textsc{agent} and $Y$ is an affected \textsc{theme}. Depending on the sense, the theme may be a body part or garment, as in rolling up sleeves, a concrete object such as bedding or laundry, a tool or net that is removed, or an activity that is brought to an end.

\begin{exe}
\ex \label{geotda-3}
    \glll $\{X=N_0\}$-이 $\{Y=N_1\}$-을 $V$\\
          ${\color{white}\{X=N_0\}}$-\textit{i} ${\color{white}\{Y=N_1\}}$-\textit{eul} \\
          ${\color{white}\{X=N_0\}}$-\textsc{nom} ${\color{white}\{Y=N_1\}}$-\textsc{acc} \\
\end{exe}

\noindent In the collection sense, the regular entry further licenses a source argument, as in (\ref{geotda-4}). Here $X$ is an \textsc{agent}, $Y$ is the collected \textsc{theme}, and $Z$ is a \textsc{source}, typically a person or group from whom money, fees, papers, or other concrete items are collected.

\begin{exe}
\ex \label{geotda-4}
    \glll $\{X=N_0\}$-이 $\{Z=N_2\}$-에게/에게서 $\{Y=N_1\}$-을 $V$\\
          ${\color{white}\{X=N_0\}}$-\textit{i} ${\color{white}\{Z=N_2\}}$-\textit{ege/egeseo} ${\color{white}\{Y=N_1\}}$-\textit{eul} \\
          ${\color{white}\{X=N_0\}}$-\textsc{nom} ${\color{white}\{Z=N_2\}}$-\textsc{dat/src} ${\color{white}\{Y=N_1\}}$-\textsc{acc} \\
\end{exe}

\noindent The contrast between 걸어 \textit{georeo} and 걷어 \textit{geodeo} therefore corresponds to a lexical split in argument structure. The ㄷ \textit{digeut} irregular predicate is organized around an agentive mover and, where present, a path-like or abstract theme argument. The regular predicate is organized around an affected theme, with an additional source argument in the collection sense. The surface alternation is thus not determined by the phonological shape of 걷다 \textit{geotda} alone, but by which lexical entry {is selected; the associated argument frame provides evidence for that lexical choice.}

\paragraph{닫다 \textit{datda}}
The verb 닫다 \textit{datda} likewise shows a split between two realization classes. The ㄷ \textit{digeut} irregular entry is realized as 달아 \textit{dara} before vowel-initial endings and denotes self-propelled motion, as in `run', `rush', or `gallop'. Its frame is intransitive, as in (\ref{datda-1}), with $X$ realized as a nominative \textsc{agent}. The selectional restriction on $X$ is limited to animals or vehicles, reflecting the motion-oriented character of this predicate.

\begin{exe}
\ex \label{datda-1}
    \glll $\{X=N_0\}$-이 $V$\\
          ${\color{white}\{X=N_0\}}$-\textit{i} \\
          ${\color{white}\{X=N_0\}}$-\textsc{nom} \\
\end{exe}

\noindent The regular entry, by contrast, is realized as 닫아 \textit{dada} and denotes `shut' or `close'. It is a transitive predicate, as in (\ref{datda-2}), where $X$ is an \textsc{agent} and $Y$ is an affected \textsc{theme}. The theme is typically a part of an object or building, such as a door, lid, gate, or shop door, but it may also be a body-part expression such as 입 \textit{ip} (`mouth') or 말문 \textit{malmun} (`opening for speech'), yielding extensions such as `close one's mouth' or `remain silent'.

\begin{exe}
\ex \label{datda-2}
    \glll $\{X=N_0\}$-이 $\{Y=N_1\}$-을 $V$\\
          ${\color{white}\{X=N_0\}}$-\textit{i} ${\color{white}\{Y=N_1\}}$-\textit{eul} \\
          ${\color{white}\{X=N_0\}}$-\textsc{nom} ${\color{white}\{Y=N_1\}}$-\textsc{acc} \\
\end{exe}

\noindent The contrast between 달아 \textit{dara} and 닫아 \textit{dada} therefore follows the same logic as the contrast in 걷다 \textit{geotda}. The ㄷ \textit{digeut} irregular predicate is organized around a single moving \textsc{agent}, while the regular predicate is organized around an \textsc{agent} acting on an affected \textsc{theme}. The realization contrast is thus aligned with a lexical opposition between self-propelled motion and caused closure, rather than with the phonological shape of 닫다 \textit{datda} alone. {The motion and closure frames help distinguish the two lexemes, and the selected lexeme determines whether ㄷ \textit{digeut} irregular realization applies.}

\paragraph{묻다 \textit{mutda}}
The verb 묻다 \textit{mutda} shows a richer split than 걷다 \textit{geotda} and 닫다 \textit{datda}, because both the regular and the ㄷ \textit{digeut} irregular entries license more than one frame. The ㄷ \textit{digeut} irregular entry is realized as 물어 \textit{mureo} before vowel-initial endings and denotes a communicative or interactional predicate, as in `ask', `question', or `hold responsible'. In its nominal-object frame, shown in (\ref{mutda-1}), $X$ is an \textsc{agent}, $Z$ is a \textsc{goal}, and $Y$ is a \textsc{theme}, typically the matter asked about or the responsibility attributed to the addressee.

\begin{exe}
\ex \label{mutda-1}
    \glll $\{X=N_0\}$-이 $\{Z=N_2\}$-에/에게 $\{Y=N_1\}$-을 $V$\\
          ${\color{white}\{X=N_0\}}$-\textit{i} ${\color{white}\{Z=N_2\}}$-\textit{e/ege} ${\color{white}\{Y=N_1\}}$-\textit{eul} \\
          ${\color{white}\{X=N_0\}}$-\textsc{nom} ${\color{white}\{Z=N_2\}}$-\textsc{loc/dat} ${\color{white}\{Y=N_1\}}$-\textsc{acc} \\
\end{exe}

\noindent The same irregular entry also takes clausal content. In (\ref{mutda-2}), the embedded clause marked by 지 \textit{ji} functions as the questioned \textsc{theme}; in (\ref{mutda-3}), the quoted clause marked by 고 \textit{go} functions as \textsc{content}. In both frames, $X$ remains the asking \textsc{agent} and $Z$ the addressee or \textsc{goal}.

\begin{exe}
\ex \label{mutda-2}
    \glll $\{X=N_0\}$-이 $\{Z=N_2\}$-에/에게 $\{Y=S\text{지}_{1}\}$-을 $V$\\
          ${\color{white}\{X=N_0\}}$-\textit{i} ${\color{white}\{Z=N_2\}}$-\textit{e/ege} ${\color{white}\{Y=S\text{지}_{1}\}}$-\textit{eul} \\
          ${\color{white}\{X=N_0\}}$-\textsc{nom} ${\color{white}\{Z=N_2\}}$-\textsc{loc/dat} ${\color{white}\{Y=S\text{지}_{1}\}}$-\textsc{acc} \\
\end{exe}

\begin{exe}
\ex \label{mutda-3}
    \glll $\{X=N_0\}$-이 $\{Z=N_2\}$-에/에게 $\{Y=S_1\}$-고 $V$\\
          ${\color{white}\{X=N_0\}}$-\textit{i} ${\color{white}\{Z=N_2\}}$-\textit{e/ege} ${\color{white}\{Y=S_1\}}$-\textit{go} \\
          ${\color{white}\{X=N_0\}}$-\textsc{nom} ${\color{white}\{Z=N_2\}}$-\textsc{loc/dat} ${\color{white}\{Y=S_1\}}$-\textsc{comp} \\
\end{exe}

\noindent The regular entry, by contrast, is realized as 묻어 \textit{mudeo}. Its central transitive-locative frame denotes placement or concealment, as in `bury', `put into', `sink into', `post', or related extensions. As shown in (\ref{mutda-4}), $X$ is an \textsc{agent}, $Y$ is a \textsc{theme}, and $Z$ is a \textsc{location}.

\begin{exe}
\ex \label{mutda-4}
    \glll $\{X=N_0\}$-이 $\{Y=N_1\}$-을 $\{Z=N_2\}$-에 $V$\\
          ${\color{white}\{X=N_0\}}$-\textit{i} ${\color{white}\{Y=N_1\}}$-\textit{eul} ${\color{white}\{Z=N_2\}}$-\textit{e} \\
          ${\color{white}\{X=N_0\}}$-\textsc{nom} ${\color{white}\{Y=N_1\}}$-\textsc{acc} ${\color{white}\{Z=N_2\}}$-\textsc{loc} \\
\end{exe}

\noindent This frame covers both concrete placement, such as burying a body or money in the ground, and extended uses such as placing an agent, troop, or abstract matter in a location. A reduced transitive frame is also available in more abstract uses, as in (\ref{mutda-5}), where $X$ is an \textsc{agent} and $Y$ is an abstract \textsc{theme}, as in forgetting, suppressing, or putting aside a past event.

\begin{exe}
\ex \label{mutda-5}
    \glll $\{X=N_0\}$-이 $\{Y=N_1\}$-을 $V$\\
          ${\color{white}\{X=N_0\}}$-\textit{i} ${\color{white}\{Y=N_1\}}$-\textit{eul} \\
          ${\color{white}\{X=N_0\}}$-\textsc{nom} ${\color{white}\{Y=N_1\}}$-\textsc{acc} \\
\end{exe}

\noindent The contrast between 물어 \textit{mureo} and 묻어 \textit{mudeo} is therefore not reducible to the segmental shape of 묻다 \textit{mutda}. The ㄷ \textit{digeut} irregular predicate is organized around an interactional frame involving an \textsc{agent}, a \textsc{goal}, and a \textsc{theme} or \textsc{content}. The regular predicate is organized around placement, concealment, or suppression, with an affected \textsc{theme} and, in its central frame, a \textsc{location}. The realization class is thus tied to lexical identity and argument structure. {More precisely, the interactional and placement frames support the identification of different lexemes, and the realization class follows from the lexeme selected.}

These three pairs show that the regular versus ㄷ \textit{digeut} irregular contrast is systematically associated with lexical contrasts in argument structure. The regular predicates are organized around an affected \textsc{theme}, with an additional \textsc{location} or \textsc{source} where the lexical meaning requires it. The ㄷ \textit{digeut} irregular predicates, by contrast, are either motion predicates organized around an agentive mover or interactional predicates organized around an \textsc{agent}, a \textsc{goal}, and a \textsc{theme} or \textsc{content}. {These argument-structural patterns are therefore diagnostic of lexical identity, rather than independent determinants of the surface alternation.} Table~\ref{tab:regular_d_irregular_pairs} summarizes these contrasts in terms of event type and argument structure.

\begin{table}[!ht]
\centering
\footnotesize
\renewcommand{\arraystretch}{1.15}
\begin{tabular}{llll}
\toprule
\textbf{Predicate} & \textbf{Regular predicate} & \textbf{ㄷ  irregular predicate} & \textbf{Lexical contrast} \\
\midrule
걷다 \textit{geotda} & manipulation, collection & motion & affected theme vs. path-like argument \\
닫다 \textit{datda} & caused closure & self-propelled motion & affected theme vs. agentive mover \\
묻다 \textit{mutda} & placement, concealment & communication & location/theme vs. goal/content \\
\bottomrule
\end{tabular}
\caption{Lexical contrasts underlying regular and ㄷ \textit{digeut} irregular realization}
\label{tab:regular_d_irregular_pairs}
\end{table}

The difference in surface realization is therefore aligned with lexical identity and subcategorization, not merely with the phonological shape of the stem. These contrasts support the broader claim that surface realization in such pairs is determined through lexical identification rather than by stem shape alone. {The pathway is thus not semantic role pattern directly to irregular form, but argument-structural evidence to lexical identity, lexical identity to realization class, and realization class to surface output.}

\subsection{Regular versus ㅂ \textit{bieup} irregular pairs}

Regular versus ㅂ \textit{bieup} irregular pairs show the same general pattern from a different inflectional class. The relevant predicates have the same or nearly the same stem shape, but their surface forms diverge according to lexical identity. In 굽다 \textit{gupda}, the contrast between 구워 \textit{guwo} and 굽어 \textit{gubeo} corresponds to a difference between a transitive externally caused event and an intransitive change-of-state predicate. In 곱다 \textit{gopda}, the contrast is less clearly reflected in argument number, but still depends on lexical differentiation: the ㅂ \textit{bieup} irregular adjective 고와 \textit{gowa} belongs to a positive property predicate, while the regular counterpart belongs to a distinct evaluative predicate. {In both cases, lexical meaning helps distinguish the intended predicate, and the selected predicate determines the realization class.} These cases show that ㅂ \textit{bieup} irregularity cannot be predicted from the final consonant of the stem alone.

\paragraph{굽다 \textit{gupda}}
The verb 굽다 \textit{gupda} shows a split between two lexical entries sharing the same headword but belonging to different realization classes. The ㅂ \textit{bieup} irregular entry is realized as 구워 \textit{guwo} before vowel-initial endings and denotes externally caused resultative or creative events, including `roast', `bake', `burn', `fire', `produce', and `copy'. These senses share the transitive frame in (\ref{gupda-1}), where $X$ is an \textsc{agent} and $Y$ is an affected or created \textsc{theme}.

\begin{exe}
\ex \label{gupda-1}
    \glll $\{X=N_0\}$-이 $\{Y=N_1\}$-을 $V$\\
          ${\color{white}\{X=N_0\}}$-\textit{i} ${\color{white}\{Y=N_1\}}$-\textit{eul} \\
          ${\color{white}\{X=N_0\}}$-\textsc{nom} ${\color{white}\{Y=N_1\}}$-\textsc{acc} \\
\end{exe}

\noindent The theme varies with the lexical sense: it may be food in the cooking sense, an artificial object such as charcoal, pottery, or bricks in the firing sense, salt in the production sense, or digital media in the copying sense. Across these uses, however, the predicate remains organized around an external \textsc{agent} bringing about a result in an affected or created \textsc{theme}.

The regular entry, by contrast, is realized as 굽어 \textit{gubeo} and denotes an internally unfolding change of shape, as in `curve' or `bend'. Its frame is intransitive, as in (\ref{gupda-2}), with $X$ realized as a nominative \textsc{theme} rather than as an agent.

\begin{exe}
\ex \label{gupda-2}
    \glll $\{X=N_0\}$-이 $V$\\
          ${\color{white}\{X=N_0\}}$-\textit{i} \\
          ${\color{white}\{X=N_0\}}$-\textsc{nom} \\
\end{exe}

\noindent In this frame, $X$ is the entity that undergoes the change of shape, such as a concrete object, a tree trunk, or a body part such as a back, waist, hand, or finger. The contrast between 구워 \textit{guwo} and 굽어 \textit{gubeo} is therefore not a difference between two phonological treatments of the same predicate. It is a lexical contrast between a ㅂ \textit{bieup} irregular transitive predicate of caused result and a regular intransitive predicate of shape change. {The transitive caused-result frame and the intransitive shape-change frame help identify distinct lexical entries; the realization contrast follows from the realization class associated with each entry.}

\paragraph{곱다 \textit{gopda}}
The adjective 곱다 \textit{gopda} differs from 굽다 \textit{gupda} in that the contrast is not primarily expressed through argument number. The ㅂ \textit{bieup} irregular entry is realized as 고와 \textit{gowa} before vowel-initial endings and denotes positive or descriptive properties such as `be beautiful', `be fine-grained', `be fine-textured', `be sweet in manner', and `be fair'. These senses share the one-place adjectival frame in (\ref{gopda-1}), where $X$ is a nominative \textsc{theme} bearing the relevant property.

\begin{exe}
\ex \label{gopda-1}
    \glll $\{X=N_0\}$-이 $A$\\
          ${\color{white}\{X=N_0\}}$-\textit{i} \\
          ${\color{white}\{X=N_0\}}$-\textsc{nom} \\
\end{exe}

\noindent The thematic argument may be a concrete object or appearance in the visual sense, a powder or textile in the fine-grained or fine-textured sense, a manner-related noun such as 마음씨 \textit{maeumssi} (`character') or 말씨 \textit{malssi} (`way of speaking') in the interpersonal sense, or a body-part expression such as 피부 \textit{pibu} (`skin') in the fairness sense. Across these uses, the adjective attributes a positive or descriptive property to a single \textsc{theme}.

The regular counterpart, realized without ㅂ \textit{bieup} irregular alternation, belongs to a distinct lexical adjective meaning roughly `turn out unfavorably' or `go badly'. Its frame is likewise one-place, as in (\ref{gopda-2}), with $X$ interpreted as the situation, matter, or course of events whose outcome is evaluated.

\begin{exe}
\ex \label{gopda-2}
    \glll $\{X=N_0\}$-이 $A$\\
          ${\color{white}\{X=N_0\}}$-\textit{i} \\
          ${\color{white}\{X=N_0\}}$-\textsc{nom} \\
\end{exe}

\noindent The contrast between 고와 \textit{gowa} and 곱아 \textit{goba} therefore cannot be explained by a difference in overt valency. Both entries are one-place adjectival predicates. What distinguishes them is lexical meaning and realization class: the ㅂ \textit{bieup} irregular entry attributes a positive or descriptive property to a theme, whereas the regular entry evaluates the unfavorable development of a situation. In this pair, lexical semantic differentiation, rather than argument-number contrast, {supports the identification of the relevant lexical entry, and the lexical entry determines} the surface realization.

These two pairs show that regular versus ㅂ \textit{bieup} irregular realization is associated with lexical contrasts, though the relevant contrast may be expressed in different grammatical ways. In 굽다 \textit{gupda}, the opposition is argument-structural: the ㅂ \textit{bieup} irregular predicate is transitive and agentive, while the regular predicate is intransitive and theme-oriented. In 곱다 \textit{gopda}, both entries are one-place adjectival predicates, but they differ in lexical meaning and evaluative orientation. {The point is therefore not that argument number or semantic class directly triggers ㅂ \textit{bieup} irregularity, but that these properties help distinguish lexemes with different realization classes.} Table~\ref{tab:regular_b_irregular_pairs} summarizes these contrasts.

\begin{table}[!ht]
\centering
\footnotesize
\begin{tabular}{llll}
\toprule
\textbf{Predicate} & \textbf{Regular predicate} & \textbf{ㅂ irregular predicate} & \textbf{Lexical contrast} \\
\midrule
굽다 \textit{gupda} & shape change & caused result, creation & theme-only change vs. agentive causation \\
곱다 \textit{gopda} & unfavorable outcome & positive property & situational evaluation vs. attributed property \\
\bottomrule
\end{tabular}
\caption{Lexical contrasts underlying regular and ㅂ \textit{bieup} irregular realization}
\label{tab:regular_b_irregular_pairs}
\end{table}

The difference in surface realization is therefore again tied to lexical identification. The final ㅂ \textit{bieup} of the stem marks the formal site where alternation may occur, but it does not determine whether the irregular alternation is selected. {As in the ㄷ \textit{digeut} pairs, the explanatory path is from lexical-semantic and argument-structural evidence to lexical identity, from lexical identity to realization class, and from realization class to surface form.}

\subsection{르 \textit{reu} irregular versus 러 \textit{reo} irregular pairs}

르 \textit{reu} irregular versus 러 \textit{reo} irregular pairs show that homonymous or near-homonymous predicates may diverge not only between regular and irregular realization, but also between two distinct irregular classes. In 누르다 \textit{nureuda}, the contrast is between an eventive transitive predicate realized as 눌러 \textit{nulleo} and a stative property predicate realized as 누르러 \textit{nureureo}. In 이르다 \textit{ireuda}, the contrast is between a resultative or directional predicate realized as 이르러 \textit{ireureo} and a communicative predicate realized as 일러 \textit{illeo}. The relevant distinction is therefore not the presence of stem-final 르 \textit{reu} alone, but the lexical entry to which the stem form belongs. {This type is especially revealing because the ambiguity is not between regular and irregular realization, but between two lexically specified irregular realization classes.}

\paragraph{누르다 \textit{nureuda}}
The predicate 누르다 \textit{nureuda} shows a split between two realization classes. The 르 \textit{reu} irregular entry is realized as 눌러 \textit{nulleo} before vowel-initial endings and denotes an eventive transitive predicate, including physical pressing, overcoming, and suppressing. These senses share the frame in (\ref{nureuda-1}), where $X$ is the force-exerting participant and $Y$ is the affected \textsc{theme}.

\begin{exe}
\ex \label{nureuda-1}
    \glll $\{X=N_0\}$-이 $\{Y=N_1\}$-을 $V$\\
          ${\color{white}\{X=N_0\}}$-\textit{i} ${\color{white}\{Y=N_1\}}$-\textit{eul} \\
          ${\color{white}\{X=N_0\}}$-\textsc{nom} ${\color{white}\{Y=N_1\}}$-\textsc{acc} \\
\end{exe}

\noindent In the physical and competitive senses, $X$ is an \textsc{agent}; in the suppressive sense, $X$ is closer to a controlling or managing participant, while $Y$ remains a \textsc{theme}, such as an emotion, desire, or excitement. Across these uses, the predicate profiles an event in which force, dominance, or control is exerted over an affected object or abstract state.

The 러 \textit{reo} irregular entry, by contrast, is realized as 누르러 \textit{nureureo} and denotes a color property, roughly `be yellowish' or `have a dull golden color'. Its frame is one-place adjectival, as in (\ref{nureuda-2}), with $X$ interpreted as the \textsc{theme} bearing the relevant property.

\begin{exe}
\ex \label{nureuda-2}
    \glll $\{X=N_0\}$-이 $A$\\
          ${\color{white}\{X=N_0\}}$-\textit{i} \\
          ${\color{white}\{X=N_0\}}$-\textsc{nom} \\
\end{exe}

\noindent The contrast between 눌러 \textit{nulleo} and 누르러 \textit{nureureo} is therefore a contrast between a transitive event predicate and a one-place property predicate. The realization class is tied to lexical meaning and argument structure, not to the segmental shape of 누르다 \textit{nureuda} alone. {More precisely, the eventive transitive frame and the stative property frame help distinguish two lexical entries, and each lexical entry is associated with a different irregular realization class.}

\paragraph{이르다 \textit{ireuda}}
The predicate 이르다 \textit{ireuda} shows a more complex split, because both realization classes license several closely related senses. The 러 \textit{reo} irregular entry is realized as 이르러 \textit{ireureo} and denotes reaching, arriving, amounting to, or coming to a result. Its core frame is shown in (\ref{ireuda-1}), where $X$ is a \textsc{theme} and $Y$ is a \textsc{goal} or final state.

\begin{exe}
\ex \label{ireuda-1}
    \glll $\{X=N_0\}$-이 $\{Y=N_1\}$-에 $V$\\
          ${\color{white}\{X=N_0\}}$-\textit{i} ${\color{white}\{Y=N_1\}}$-\textit{e} \\
          ${\color{white}\{X=N_0\}}$-\textsc{nom} ${\color{white}\{Y=N_1\}}$-\textsc{loc} \\
\end{exe}

\noindent In concrete arrival uses, $Y$ is a spatial \textsc{goal}; in temporal uses, it is a time or age endpoint; in abstract resultative uses, it is a final state or amount. The subject is therefore not an agentive speaker, but an entity that reaches a place, time, state, or quantity.

The 르 \textit{reu} irregular entry, by contrast, is realized as 일러 \textit{illeo} and belongs to a communicative semantic class, including `advise', `admonish', `tell', `inform', `report', and `name'. In clausal-content frames, as in (\ref{ireuda-2}) and (\ref{ireuda-3}), $X$ is an \textsc{agent}, $Z$ is an addressee-like \textsc{goal}, and $Y$ is the communicated \textsc{content} or intended result.

\begin{exe}
\ex \label{ireuda-2}
    \glll $\{X=N_0\}$-이 $\{Z=N_2\}$-에게 $\{Y=S_1\}$-고 $V$\\
          ${\color{white}\{X=N_0\}}$-\textit{i} ${\color{white}\{Z=N_2\}}$-\textit{ege} ${\color{white}\{Y=S_1\}}$-\textit{go} \\
          ${\color{white}\{X=N_0\}}$-\textsc{nom} ${\color{white}\{Z=N_2\}}$-\textsc{dat} ${\color{white}\{Y=S_1\}}$-\textsc{comp} \\
\end{exe}

\begin{exe}
\ex \label{ireuda-3}
    \glll $\{X=N_0\}$-이 $\{Z=N_2\}$-에게 $\{Y=S_1\}$-도록 $V$\\
          ${\color{white}\{X=N_0\}}$-\textit{i} ${\color{white}\{Z=N_2\}}$-\textit{ege} ${\color{white}\{Y=S_1\}}$-\textit{dorok} \\
          ${\color{white}\{X=N_0\}}$-\textsc{nom} ${\color{white}\{Z=N_2\}}$-\textsc{dat} ${\color{white}\{Y=S_1\}}$-\textsc{comp} \\
\end{exe}

\noindent The same communicative entry also licenses nominal-content frames, as in (\ref{ireuda-4}), where $Y$ is a \textsc{theme}, and naming frames, as in (\ref{ireuda-5}), where $Z$ functions as a resultative label or final state.

\begin{exe}
\ex \label{ireuda-4}
    \glll $\{X=N_0\}$-이 $\{Z=N_2\}$-에게 $\{Y=N_1\}$-을 $V$\\
          ${\color{white}\{X=N_0\}}$-\textit{i} ${\color{white}\{Z=N_2\}}$-\textit{ege} ${\color{white}\{Y=N_1\}}$-\textit{eul} \\
          ${\color{white}\{X=N_0\}}$-\textsc{nom} ${\color{white}\{Z=N_2\}}$-\textsc{dat} ${\color{white}\{Y=N_1\}}$-\textsc{acc} \\
\end{exe}

\begin{exe}
\ex \label{ireuda-5}
    \glll $\{X=N_0\}$-이 $\{Y=N_1\}$-을 $\{Z=N_2\}$-라고 $V$\\
          ${\color{white}\{X=N_0\}}$-\textit{i} ${\color{white}\{Y=N_1\}}$-\textit{eul} ${\color{white}\{Z=N_2\}}$-\textit{rago} \\
          ${\color{white}\{X=N_0\}}$-\textsc{nom} ${\color{white}\{Y=N_1\}}$-\textsc{acc} ${\color{white}\{Z=N_2\}}$-\textsc{quot} \\
\end{exe}

\noindent The contrast between 이르러 \textit{ireureo} and 일러 \textit{illeo} is therefore a contrast between two lexical predicates with different argument structures. The 러 \textit{reo} irregular predicate is organized around a \textsc{theme} reaching a \textsc{goal} or final state, while the 르 \textit{reu} irregular predicate is organized around an \textsc{agent}, a communicated \textsc{theme} or \textsc{content}, and an addressee-like \textsc{goal}. {These frames help identify the resultative or communicative lexeme, and the identified lexeme determines whether 러 \textit{reo} or 르 \textit{reu} irregular realization is selected.}

These two pairs show that the 르 \textit{reu} irregular versus 러 \textit{reo} irregular contrast is systematically associated with lexical contrasts in event type and argument structure. In 누르다 \textit{nureuda}, the opposition is between a transitive action predicate and a one-place property predicate. In 이르다 \textit{ireuda}, the opposition is between a resultative or directional predicate and a communicative predicate. {These semantic and argument-structural contrasts are diagnostic of lexical identity rather than direct triggers of either irregular pattern.} Table~\ref{tab:reu_reo_irregular_pairs} summarizes these contrasts.

\begin{table}[!ht]
\centering
\footnotesize
\begin{tabular}{llll}
\toprule
\textbf{Predicate} & \textbf{르 irregular predicate} & \textbf{러 irregular predicate} & \textbf{Lexical contrast} \\
\midrule
누르다 \textit{nureuda} & transitive action, control & color property & agentive force vs. attributed property \\
이르다 \textit{ireuda} & communication, naming & reaching, result & content transfer vs. goal attainment \\
\bottomrule
\end{tabular}
\caption{Lexical contrasts underlying 르 \textit{reu} irregular and 러 \textit{reo} irregular realization}
\label{tab:reu_reo_irregular_pairs}
\end{table}

The choice between 르 \textit{reu} irregular and 러 \textit{reo} irregular realization is therefore determined through lexical identification. Stem-final 르 \textit{reu} marks the formal site of the alternation, but it does not determine which irregular realization is selected. {The explanatory sequence is again lexical-semantic and argument-structural evidence to lexical identity, lexical identity to realization class, and realization class to surface form.}

\subsection{Scope of the inventory}
The inventory presented in this section delimits the simplex Korean predicates that instantiate the realization ambiguity examined in this paper. The relevant criterion is not irregularity alone, but homonymy with inflectional divergence: lexical items with the same or nearly the same stem shape are assigned to different realization classes, so that a canonical stem-ending representation does not by itself determine a unique surface form. The cases discussed above instantiate three such contrasts: regular versus ㄷ \textit{digeut} irregular realization, regular versus ㅂ \textit{bieup} irregular realization, and 르 \textit{reu} irregular versus 러 \textit{reo} irregular realization. The inventory is therefore not a general catalogue of Korean irregular predicates, but a focused set of predicates in which surface realization is underdetermined by canonical morpheme sequence alone. {What unifies the inventory is the need to identify the intended lexeme before the appropriate realization class can be selected.}
Further examples arise through derived predicates built on the same lexical bases, such as 치걷다 \textit{chigeotda} from 걷다 \textit{geotda}, and 되묻다 \textit{doemutda} and 파묻다 \textit{pamutda} from 묻다 \textit{mutda}. These derived predicates do not introduce independent ambiguity classes. Rather, they inherit the realization behavior of the simplex base and show that the lexical contrast remains stable when the relevant base appears inside a larger predicate. {They are therefore important as extensions of the same realization classes, but they do not change the core inventory of simplex ambiguity types.} For this reason, they are treated as extensions of the simplex patterns rather than as separate members of the core inventory.

\section{Implications for the description of Korean morphology}

The patterns examined in this paper bear directly on how Korean morphology should be described. The first implication is that morphological analysis and surface realization cannot be treated as simple inverses of one another. A surface form may be segmented into an ordered sequence of morphemes and grammatical categories, yet that same representation may fail to determine a unique output when read in the opposite direction. The descriptive adequacy of morphological analysis in decomposing an eojeol does not therefore guarantee descriptive adequacy for realization. These are related but distinct grammatical relations, and they require different kinds of information.

This asymmetry matters because Korean morphology is often discussed as though the output of analysis were little more than the surface string rendered in segmented form. The cases considered here show that such a view is insufficient. A morphologically analyzed representation is an abstraction over surface form, not a reversible encoding of it. Once realization is approached as an independent problem, it becomes clear that the grammar must specify not only what morphemes are present, but also which lexical item and realization class are involved. In this sense, surface realization provides a test of representational sufficiency. It reveals where segmentation and category labels determine the overt form, where they approximate it, and where they leave crucial distinctions unresolved. {Homonymy with inflectional divergence marks the strongest version of this unresolved relation, since the analyzed stem-ending configuration remains compatible with more than one surface form until lexical identity is fixed.}

A second implication concerns the description of irregular conjugation. Korean irregularity is not adequately characterized if it is presented only as a set of phonological alternations triggered by local environments. Such descriptions capture an important part of the pattern, but they presuppose that the relevant lexical class has already been identified. The cases examined here show that predicates with the same or nearly the same segmental shape may belong to different realization classes, and that this difference may correlate systematically with lexical meaning. Irregularity is therefore not simply a matter of stem shape plus following ending. It is a property of lexical items as members of particular realization classes.

From this perspective, the traditional opposition between regular and irregular forms is descriptively too coarse when applied to the ambiguity cases discussed here. What is required is a more articulated account in which realization patterns are distributed across lexically specified classes whose members behave systematically once identified. The problem is not that Korean contains a residue of accidental exceptions beyond the reach of grammar. It is that the grammar must encode the fact that realization is mediated by lexical classification. A purely form-based treatment obscures this point because it treats phonological shape as if it were sufficient to determine class membership. The ambiguity pairs examined here show that it is not. {Their significance lies precisely in the fact that formal shape identifies the possible site of alternation, but not the realization class itself.}

The argument also bears on the relation between morphology and lexical semantics. In the strongest cases, the contrast in surface realization does not follow merely from formal class membership, but from the fact that different lexical meanings associated with the same stem shape are assigned to different realization classes. The relevant distinction is therefore not external to morphology in the sense of being accidental or post hoc. It is part of the lexical information on which realization depends. This does not mean that morphological realization should be reduced to semantics, nor that semantic roles replace morphophonological description. It means that a full description of Korean predicate realization cannot sharply isolate realization from lexical meaning when the choice of surface form depends on which predicate is intended. {The relevant dependency is indirect: lexical meaning helps identify the predicate, the predicate carries a realization class, and that realization class determines the surface output.}

The same point extends to argument structure. Where homonymous predicates differ not only in meaning but also in the semantic roles associated with their arguments, those distinctions provide a principled basis for lexical identification. Surface realization is thus connected not only to lexical storage and morphophonological patterning, but also to the way predicates structure their participants. The architecture suggested by these cases is one in which morphology, lexical semantics, and argument structure interact at the point where a lexical item is selected for realization. Korean predicate morphology makes this interaction visible because the relevant lexical distinctions surface directly in inflectional form. {This interaction does not make argument structure a rule of realization; rather, argument structure is part of the evidence by which the intended lexeme is distinguished.}

For the description of Korean grammar more generally, this means that surface realization should be treated as a legitimate object of linguistic analysis in its own right. It is not merely a practical issue of reconstructing forms from segmented input, nor only a pedagogical matter concerning irregular paradigms. It is a domain in which the internal organization of the lexicon, the structure of inflectional classes, and the interaction between morphology and meaning become empirically accessible. By asking what information is required for a morphologically analyzed form to determine a unique surface output, one gains a sharper view of what Korean morphology stores, what it derives, and where its descriptive categories must be enriched.

The broader significance of the present argument lies in the perspective it offers on Korean grammar itself. Surface realization shows that the passage from structure to form is not exhausted by linear recombination or local phonological adjustment. It depends on lexical identity, realization class membership, and, in the ambiguity cases examined here, lexical-semantic and argument-structural evidence for distinguishing the intended predicate. A description of Korean morphology that omits these factors remains incomplete not because it fails to list enough alternations, but because it misses the conditions under which alternation becomes grammatically determinate. In this respect, surface realization is not peripheral to the grammar. It is one of the points at which the grammar's internal organization becomes most clearly visible.

\section{Conclusion}

This paper has examined Korean surface realization from the perspective of the inverse relation between morphological analysis and overt form. Starting from a sequence of canonical morphemes and grammatical category labels, it asked under what conditions such a representation determines a unique surface output. The answer developed here is that it does not do so in any general sense. While many forms, especially in the nominal domain, approach transparent realization, predicate morphology reveals a more complex interaction among stem shape, lexical identity, realization class membership, and morphophonological conditioning.

The discussion has shown that Korean irregular conjugation cannot be understood adequately as a simple inventory of local phonological alternations. Such alternations describe what happens once a predicate has been assigned to the relevant realization class. They do not, by themselves, determine how that class is identified. Predicates with identical or near-identical stem shapes may belong to different realization classes, and the resulting surface forms may diverge in the same formal environment. {The strongest cases are therefore not merely cases of homonymy, nor merely cases of irregular conjugation, but cases of homonymy with inflectional divergence: distinct lexical items associated with the same or nearly the same stem shape are assigned to different realization classes and yield different surface forms.}

On this basis, the paper has argued that Korean surface realization is underdetermined by bare morphological representation in a restricted but theoretically significant set of cases. The ambiguity is not accidental, nor does it consist only of isolated lexical curiosities. It is organized into recurrent classes, including regular versus ㄷ \textit{digeut} irregular, regular versus ㅂ \textit{bieup} irregular, and 르 \textit{reu} irregular versus 러 \textit{reo} irregular contrasts. A central contribution of the present study has been to identify these classes as a coherent problem for surface realization, rather than treating them as scattered observations within the broader description of irregular conjugation.

The paper has further argued that the relevant disambiguating information is not purely morphological, but neither is realization directly determined by semantic roles. In the clearest cases, lexical meaning, subcategorization, and semantic role structure help identify the intended predicate. The predicate, as a lexical item, determines the realization class, and the realization class determines the surface output. {The explanatory path is therefore lexical-semantic and argument-structural evidence to lexical identity, lexical identity to realization class, and realization class to surface form.} This preserves a morphology-first account of realization while showing why lexical and argument-structural information is necessary for resolving the ambiguity.

{This analysis also motivates Korean surface realization as a concrete language-processing problem. If a morphologically analyzed representation can remain compatible with more than one surface form, then reconstruction from morpheme and part-of-speech input cannot be treated as simple concatenation or local spelling normalization. It requires access to realization classes and, in the ambiguity cases, to lexical identity in context. The corpus counts and reconstruction results reported in the appendix support this view: the relevant predicate families are attested in KLUE, and surface reconstruction over morphologically analyzed input improves substantially when explicit realization constraints are modeled. These results do not replace the linguistic analysis, but they show that the theoretical distinction has practical consequences for Korean morphological processing.}

For this reason, Korean surface realization deserves to be treated as a linguistic problem in its own right. It offers a precise domain in which to examine what morphological representation contains, what it leaves unspecified, and what additional information is required for form to become uniquely determined. The study of ambiguous predicate realization thus clarifies more than a limited set of verbal and adjectival alternations. It sheds light on the organization of Korean morphology itself, and on the broader question of how lexical identity, inflectional class membership, and grammatical form are connected in the language.


\begin{thebibliography}{12}
\providecommand{\natexlab}[1]{#1}
\providecommand{\url}[1]{\texttt{#1}}
\expandafter\ifx\csname urlstyle\endcsname\relax
  \providecommand{\doi}[1]{doi: #1}\else
  \providecommand{\doi}{doi: \begingroup \urlstyle{rm}\Url}\fi

\bibitem[Anderson(1992)]{anderson-1992-a-morphous}
Stephen~R. Anderson.
\newblock \emph{{A-Morphous Morphology}}, volume~62.
\newblock Cambridge University Press, Cambridge, 1992.
\newblock ISBN 9780521378666.
\newblock \doi{DOI: 10.1017/CBO9780511586262}.
\newblock URL \url{https://www.cambridge.org/core/product/C1CA438B04929EA5B92FF84A20DF6894}.

\bibitem[Cho(2009)]{cho-2009-study}
Sungmoon Cho.
\newblock {A Study of ‘ㅸ’s Temporal Change in Optimality Theory}.
\newblock \emph{The Linguistic Association of Korea Journal}, 17\penalty0 (2):\penalty0 93--110, 2009.

\bibitem[Ha(2012)]{ha-2012-verbal}
Sekyeong Ha.
\newblock {Verbal Inflectional Paradigms and Homophony Avoidance Constraints}.
\newblock \emph{EONEOHAG}, 62\penalty0 (1):\penalty0 69--100, 2012.
\newblock ISSN 1225-7494.

\bibitem[Kim(2015{\natexlab{a}})]{kim-2015-conjugated}
Bong-Gook Kim.
\newblock {Conjugated form and phonology}.
\newblock \emph{Korean Language and Literature}, 59\penalty0 (1):\penalty0 5--22, 2015{\natexlab{a}}.
\newblock ISSN 1229-3946.
\newblock \doi{10.23016/kllj.2015.59.59.5}.

\bibitem[Kim(2015{\natexlab{b}})]{kim-2015-establishment}
Ryangjin Kim.
\newblock {The establishment of word and the concept of word-phrase}.
\newblock \emph{The Korean Language and Literature}, 171\penalty0 (1):\penalty0 5--39, 2015{\natexlab{b}}.
\newblock ISSN 0451-0097.
\newblock \doi{10.17291/kolali.2015..171.001}.

\bibitem[Lim(2024)]{lim-2024-terms}
Dong-Hoon Lim.
\newblock {Terms in Korean School Grammar from the Perspective of Theoretical Grammar}.
\newblock \emph{Journal of Korean Linguistics}, 111\penalty0 (1):\penalty0 109--132, 2024.
\newblock ISSN 1225-1933.
\newblock \doi{10.15811/jkl.2024..111.004}.

\bibitem[Park et~al.(2021)Park, Moon, Kim, Cho, Han, Park, Song, Kim, Song, Oh, Lee, Oh, Lyu, Jeong, Lee, Seo, Lee, Kim, Lee, Jang, Do, Kim, Lim, Lee, Park, Shin, Kim, Park, Oh, Ha, and Cho]{park-etal-2021-klue}
Sungjoon Park, Jihyung Moon, Sungdong Kim, Won~Ik Cho, Ji~Yoon Han, Jangwon Park, Chisung Song, Junseong Kim, Youngsook Song, Taehwan Oh, Joohong Lee, Juhyun Oh, Sungwon Lyu, Younghoon Jeong, Inkwon Lee, Sangwoo Seo, Dongjun Lee, Hyunwoo Kim, Myeonghwa Lee, Seongbo Jang, Seungwon Do, Sunkyoung Kim, Kyungtae Lim, Jongwon Lee, Kyumin Park, Jamin Shin, Seonghyun Kim, Lucy Park, Alice Oh, Jung-Woo Ha, and Kyunghyun Cho.
\newblock {KLUE: Korean Language Understanding Evaluation}.
\newblock In Joaquin Vanschoren and Serena Yeung, editors, \emph{Proceedings of the Neural Information Processing Systems Track on Datasets and Benchmarks}, volume~1, pages 1--25. Curran, 2021.

\bibitem[Park(2004)]{park-2004-irregularity}
Sunwoo Park.
\newblock {On the irregularity of Korean unsystematic conjugation}.
\newblock \emph{Journal of CheongRam Korean Language Education}, 30\penalty0 (1):\penalty0 223--249, 2004.
\newblock ISSN 1598-1967.

\bibitem[Song(1995)]{song-1995-irregularities}
Cheol~Eui Song.
\newblock {Irregularities in Declension and Conjugation}.
\newblock \emph{Chintan Hakpo}, 80\penalty0 (1):\penalty0 273--290, 1995.
\newblock URL \url{https://kiss.kstudy.com/Detail/Ar?key=160166}.

\bibitem[Stump(2001)]{stump-2001-inflectional}
Gregory~T. Stump.
\newblock \emph{{Inflectional Morphology: A Theory of Paradigm Structure}}, volume~93.
\newblock Cambridge University Press, Cambridge, 2001.
\newblock ISBN 9780521780476.
\newblock \doi{DOI: 10.1017/CBO9780511486333}.
\newblock URL \url{https://www.cambridge.org/core/product/4A2E166618A2B0668D217AA6E057C65F}.

\bibitem[Stump(2016)]{stump-2016-inflectional}
Gregory~T. Stump.
\newblock \emph{{Inflectional Paradigms: Content and Form at the Syntax–Morphology Interface}}.
\newblock Number 149 in Cambridge Studies in Linguistics. Cambridge University Press, Cambridge, 2016.
\newblock ISBN 9781107460850.
\newblock \doi{DOI:10.1017/CBO9781316105290}.

\bibitem[Yoon et~al.(2023)Yoon, Park, Kim, Cho, Park, Kim, Seo, and Oh]{yoon-etal-2023-towards}
Soyoung Yoon, Sungjoon Park, Gyuwan Kim, Junhee Cho, Kihyo Park, Gyu~Tae Kim, Minjoon Seo, and Alice Oh.
\newblock {Towards standardizing Korean Grammatical Error Correction: Datasets and Annotation}.
\newblock In \emph{Proceedings of the 61st Annual Meeting of the Association for Computational Linguistics (Volume 1: Long Papers)}, pages 6713--6742, Toronto, Canada, 7 2023. Association for Computational Linguistics.
\newblock URL \url{https://aclanthology.org/2023.acl-long.371}.

\end{thebibliography}

\appendix

\section{Corpus counts of ambiguity predicates in KLUE}
\label{app:klue_count}

We conducted a corpus-based count of the ambiguity predicates discussed in the main text using the publicly available training and development splits of the KLUE dependency treebank \citep{park-etal-2021-klue}. The purpose of this appendix is not to estimate the productivity of Korean irregular classes, but to document whether the predicate families analyzed in the paper are attested in a contemporary annotated corpus and how their tokens are distributed across realization classes. {The counts also provide a corpus-based check on the central claim of the paper: the relevant contrasts are not merely constructed paradigm examples, but attested cases in which lexical identity matters for surface realization.}

The target set consists of predicates belonging to the three ambiguity classes examined in the paper: regular versus ㄷ \textit{digeut} irregular realization, regular versus ㅂ \textit{bieup} irregular realization, and 르 \textit{reu} irregular versus 러 \textit{reo} irregular realization. We included both simplex predicates and derived predicates whose inflectional behavior follows the corresponding base predicate. Table~\ref{tab:klue_derived_predicates} lists the predicate families included in the count.

\begin{table}[!th]
  \centering
  \footnotesize
  \begin{tabular}{p{3cm} p{9cm}}
  \toprule
  \textbf{Base predicate} & \textbf{Attested predicates included in the count} \\
  \midrule
  \{걷다\} \textit{geotda} & 걷다 \textit{geotda} \\
  \{묻다\} \textit{mutda} & 묻다 \textit{mutda} \\
  \{닫다\} \textit{datda} & 닫다 \textit{datda}, 깨닫다 \textit{kkaedatda} \\
  \{굽다\} \textit{gupda} & 굽다 \textit{gupda} \\
  \{누르다\} \textit{nureuda} & 누르다 \textit{nureuda}, 억누르다 \textit{eongnureuda}, 짓누르다 \textit{jinnureuda} \\
  \{이르다\} \textit{ireuda} & 이르다 \textit{ireuda} \\
  \bottomrule
  \end{tabular}
  \caption{Predicate families included in the KLUE count}
  \label{tab:klue_derived_predicates}
\end{table}

Each attested token was manually classified in sentential context. Classification was based on {contextual lexical identification} rather than surface form alone. A token was counted as belonging to a given realization class only when its contextual meaning corresponded to the relevant lexeme. The counting unit was the corpus token, so each occurrence was counted once regardless of inflected form. {This procedure reflects the analysis in the main text: realization class is assigned to the lexical item, and the lexical item must be identified from context.}

Table~\ref{tab:klue_ambiguity_counts} reports the distribution of regular and irregular tokens in the KLUE training and development splits.

\begin{table}[!th]
  \centering
  \footnotesize
  \begin{tabular}{l cc cc cc}
  \toprule
  & \multicolumn{2}{c}{Train} & \multicolumn{2}{c}{Dev} & \multicolumn{2}{c}{\cellcolor{gray!20}Total} \\
  Predicate family & Regular & Irregular & Regular & Irregular & \cellcolor{gray!20}Regular & \cellcolor{gray!20}Irregular \\
  \midrule
  \{걷다\} \textit{geotda} & 5 & 105 & 1 & 19 & \cellcolor{gray!20}6 & \cellcolor{gray!20}124 \\
  \{묻다\} \textit{mutda} & 6 & 19 & 0 & 0 & \cellcolor{gray!20}6 & \cellcolor{gray!20}19 \\
  \{닫다\} \textit{datda} & 4 & 13 & 2 & 0 & \cellcolor{gray!20}6 & \cellcolor{gray!20}13 \\
  \{굽다\} \textit{gupda} & 0 & 2 & 0 & 0 & \cellcolor{gray!20}0 & \cellcolor{gray!20}2 \\
  \{누르다\} \textit{nureuda} & 0 & 10 & 0 & 0 & \cellcolor{gray!20}0 & \cellcolor{gray!20}10 \\
  \{이르다\} \textit{ireuda} & 0 & 27 & 0 & 6 & \cellcolor{gray!20}0 & \cellcolor{gray!20}33 \\
  \bottomrule
  \end{tabular}
  \caption{Token counts of regular and irregular ambiguity classes in the KLUE dependency treebank}
  \label{tab:klue_ambiguity_counts}
\end{table}

The 르 \textit{reu} irregular versus 러 \textit{reo} irregular contrast was also counted separately, since both sides of this contrast are irregular realization classes. The results are given in Table~\ref{tab:klue_reu_reo_counts}.

\begin{table}[!th]
  \centering
  \footnotesize
  \begin{tabular}{l cc cc cc}
  \toprule
  & \multicolumn{2}{c}{Train} & \multicolumn{2}{c}{Dev} & \multicolumn{2}{c}{\cellcolor{gray!20}Total} \\
  Predicate family & 르 irregular & 러 irregular & 르 irregular & 러 irregular & \cellcolor{gray!20}르 irregular &
  \cellcolor{gray!20}러 irregular \\
  \midrule
  \{누르다\} \textit{nureuda} & 10 & 0 & 0 & 0 & \cellcolor{gray!20}10 & \cellcolor{gray!20}0 \\
  \{이르다\} \textit{ireuda} & 11 & 16 & 2 & 4 & \cellcolor{gray!20}13 & \cellcolor{gray!20}20 \\
  \bottomrule
  \end{tabular}
  \caption{Token counts of 르 \textit{reu} irregular and 러 \textit{reo} irregular predicates in KLUE}
  \label{tab:klue_reu_reo_counts}
\end{table}

The counts show that several ambiguity classes discussed in the main text are attested in KLUE, but with highly uneven frequency. The 걷다 \textit{geotda} family is frequent and strongly skewed toward the ㄷ \textit{digeut} irregular motion predicate, while 묻다 \textit{mutda} and 닫다 \textit{datda} show both regular and irregular tokens. The 굽다 \textit{gupda}, 누르다 \textit{nureuda}, and 이르다 \textit{ireuda} families are attested only on one side of the regular versus irregular count in Table~\ref{tab:klue_ambiguity_counts}. However, 이르다 \textit{ireuda} shows both 르 \textit{reu} irregular and 러 \textit{reo} irregular realizations when the two irregular classes are separated in Table~\ref{tab:klue_reu_reo_counts}. No token of 곱다 \textit{gopda} was found in the searched splits. {The corpus distribution therefore supports the existence of the ambiguity classes while also showing that their attestation is lexically and genre dependent.}

These counts should be interpreted as corpus evidence for attestation, not as estimates of grammatical productivity or lexical completeness. KLUE is not designed as a lexicographic resource for inflectional ambiguity, and low frequency or absence in this corpus does not imply marginality in Korean. Since the relevant distinctions are lexically conditioned, token classification required contextual interpretation and could not be recovered from morphological annotation alone. {This limitation is theoretically informative rather than merely methodological: the corpus annotation provides morpheme and part-of-speech information, but the realization contrast still requires lexical identification in context.}

As additional computational support, we evaluated surface reconstruction over the full KLUE dependency treebank. This experiment was not a direct evaluation of only the ambiguity predicates counted above, but a broader test of the inverse mapping from a sentence context and a morpheme and part-of-speech representation to a surface eojeol. {In this task, the input consists of the sentence context together with a segmented morpheme sequence and its part-of-speech labels, and the output is the corresponding orthographic surface eojeol.} 
A previous rule-based reconstruction baseline achieved a target-\textit{eojeol} $F_1$ score of 64.96 on KLUE \citep{yoon-etal-2023-towards}, while our refined rule-based system achieved a target-\textit{eojeol} $F_1$ score of 92.56 and a sentence-level accuracy of 85.59. Our system more systematically implements Korean orthographic and morphological norms (\textit{Hangugeo eomun gyubeom}) through constraints on morpheme boundaries, part-of-speech configurations, lexical stem classes, and predicate realization classes. This improvement indicates that Korean surface reconstruction requires explicit realization constraints rather than simple morpheme concatenation.\footnote{{For comparison, \texttt{GPT-5.5} achieved a target-\textit{eojeol} $F_1$ score of 73.42 and a sentence-level accuracy of 54.57 on KLUE. It outperformed the Korean-language-specific LLMs evaluated but remained well below the refined rule-based system.}}
The remaining errors, however, also show that reconstruction is not fully solved by rule-based constraints alone and may still require finer-grained lexical and contextual information, as illustrated by the ambiguity cases analyzed in the paper. {The computational task thus operationalizes the theoretical claim: morphological analysis is not always a reversible encoding of surface form, and successful reconstruction depends on information about realization classes and lexical identity.}

\end{document}